\documentclass[10pt, a4paper]{bmc_article}

\usepackage{cite} 
\usepackage{url}  
\usepackage{ifthen}  
\usepackage{multicol}   
\usepackage[latin1]{inputenc} 
\usepackage{graphicx}

\newboolean{publ}

\newenvironment{bmcformat}{\begin{raggedright}\baselineskip20pt\sloppy\setboolean{publ}{false}}{\end{raggedright}\baselineskip20pt\sloppy}

\usepackage[ngerman, USenglish, english]{babel}
\usepackage{tabularx}
\newcommand{\code}[1]{\texttt{#1}}
\def\toprule{}
\def\midrule{}
\def\bottomrule{}

\begin{document}
\nocite{GMDS2002, GMDS2003, GMDS2004}

\begin{bmcformat}

\title{Authoring case based training by document data extraction}

\author{
        Christian Betz\correspondingauthor$^1$%
        \email{Christian Betz\correspondingauthor - betz@informatik.uni-wuerzburg.de}%
      \and
        Alexander Hˆrnlein\correspondingauthor$^1$%
        \email{Alexander Hˆrnlein - hoernlein@informatik.uni-wuerzburg.de}
      and
        Frank Puppe$^1$%
        \email{Frank Puppe - puppe@informatik.uni-wuerzburg.de}%
      }

\address{%
    \iid(1)Artificial Intelligence Department, Institute of Computer Science, University of W¸rzburg,%
        Am Hubland, 97074 W¸rzburg, Germany
}%
\maketitle

\begin{abstract}
        \paragraph*{Background:} Modeling is the bottleneck to successful implementation of knowledge
management systems. In this paper, we propose an evolutionary
approach to modeling based upon word processing documents and we
describe the tool \textit{Phoenix} providing the technical
infrastructure.

        \paragraph*{Methods:} We applied our approach and software system to authoring of medical
case based training systems. So far, authors needed to either
hand-code the content (usually as HTML) or to use highly
sophisticated authoring systems which require instructions and
experience to master the complex systems. With our approach we carry
further the ideas Felciano and Dev put into practice in their system
Short Rounds \cite{Felciano1994}. They only presented pre-existing
documents as an electronic patient record. Following our approach of
evolutionary modeling, authors annotate documents to build fully
flavored diagnostic training cases \cite{Betz2003}.

        \paragraph*{Results:} For our training environment
\textit{d3web.Train}\footnote{\url{http://www.d3webtrain.de}}
\cite{Hoernlein2002, Reimer2004}, we developed a tool to extract
case knowledge from existing documents, usually dismissal records,
extending Phoenix to \textit{d3web.CaseImporter}
\cite{Hoernlein2004}. Independent authors used this tool to develop
training systems e.g. in rheumatology, gastroenterology, and
cytology, observing a significant decrease of time for setteling-in
(from several month down to 1 hour) and a decrease of time necessary
for developing a case (down to 4-6 hours) \cite{Kraemer2005}.

        \paragraph*{Conclusions:} This paper describes the general approach and provides an in-depth
analysis of the document parsing engine (Phoenix) \footnote{Phoenix
is available under LGPL open source license from
\url{https://sourceforge.net/projects/phoenix-ie/}.}. To generalize
the success of d3web.CaseImporter, we conclude by sketching further
existing applications of Phoenix, including a method to populate the
expert system d3web\footnote{\url{http://www.d3web.de}} /
Assist\footnote{\url{http://www.knowit-software.de}} and extensions
still to come (e.g. for populating the Semantic
Web\cite{w3cSemanticWeb}).
\end{abstract}

\ifthenelse{\boolean{publ}}{\begin{multicols}{2}}{}

\section{Motivation}
Phoenix is a rule-based extraction engine to transform XML documents
to arbitrary output formats. Our target developing \textit{Phoenix}
was to ``compile'' medical training cases, particularly for
\textit{d3web.Train}, from Word or Open Office documents.

By building upon well known tools and by re-using existing
documents, we seek to reduce authors' efforts both for learning and
actual modeling.

Following an evolutionary approach, authors alter and enhance
existing content (dismissal records) step-by-step to model the
desired content (training cases) \cite{Betz2004, Hoernlein2004}. For
example, an author anonymizes observations (names, locations,
dates), re-formats the document to enable automatic parsing, adds
introduction and conclusion, formulates questions and provides
feedback knowledge.

As experience shows, we succeeded in our goals to reduce learning
time and authoring effort \cite{Kraemer2005}. This led to an
increasing popularity among authors, since they can easily provide
case-based supplements to lectures.

However, Phoenix is by design a general-purpose tool and is used
e.g. for populating xml content management systems and knowledge
based systems\footnote{E.g. for \textit{Assist},
\url{http://www.knowit-software.de}}.

First, we introduce Phoenix in high-level overview, going into
details in the following sections. The second section analyzes the
Phoenix information extraction algorithm, followed by an in depth
look to the extension mechanisms necessary to process arbitrary
content and store the information into the desired format. In
addition, we show how to store the information extracted to XML back
again. After that, we describe syntax and semantics of the Phoenix
grammar, followed by the API definition. The paper closes with a
look at existing Phoenix applications apart from case authoring and
a lookout on features to come.

\section{Description}

Phoenix is a java-based engine to be extended in order to match
concrete requirements. Figure \ref{fig:structure} shows the
architecture of Phoenix. As a rule-based extraction engine, Phoenix
is initialized from a \textit{rule set}. An \textit{user object}
holds the information extracted, and together with the document is
input to Phoenix. To be applicable to different domains, Phoenix
provides two extension mechanisms: \textit{Selectors} to read
information from the document and \textit{actions} to process and
store the information to the user object.

\begin{figure}[!ht]
\centering\includegraphics[keepaspectratio=true, width=.8\textwidth, bb=0mm 0mm 171mm
91mm,clip=true]{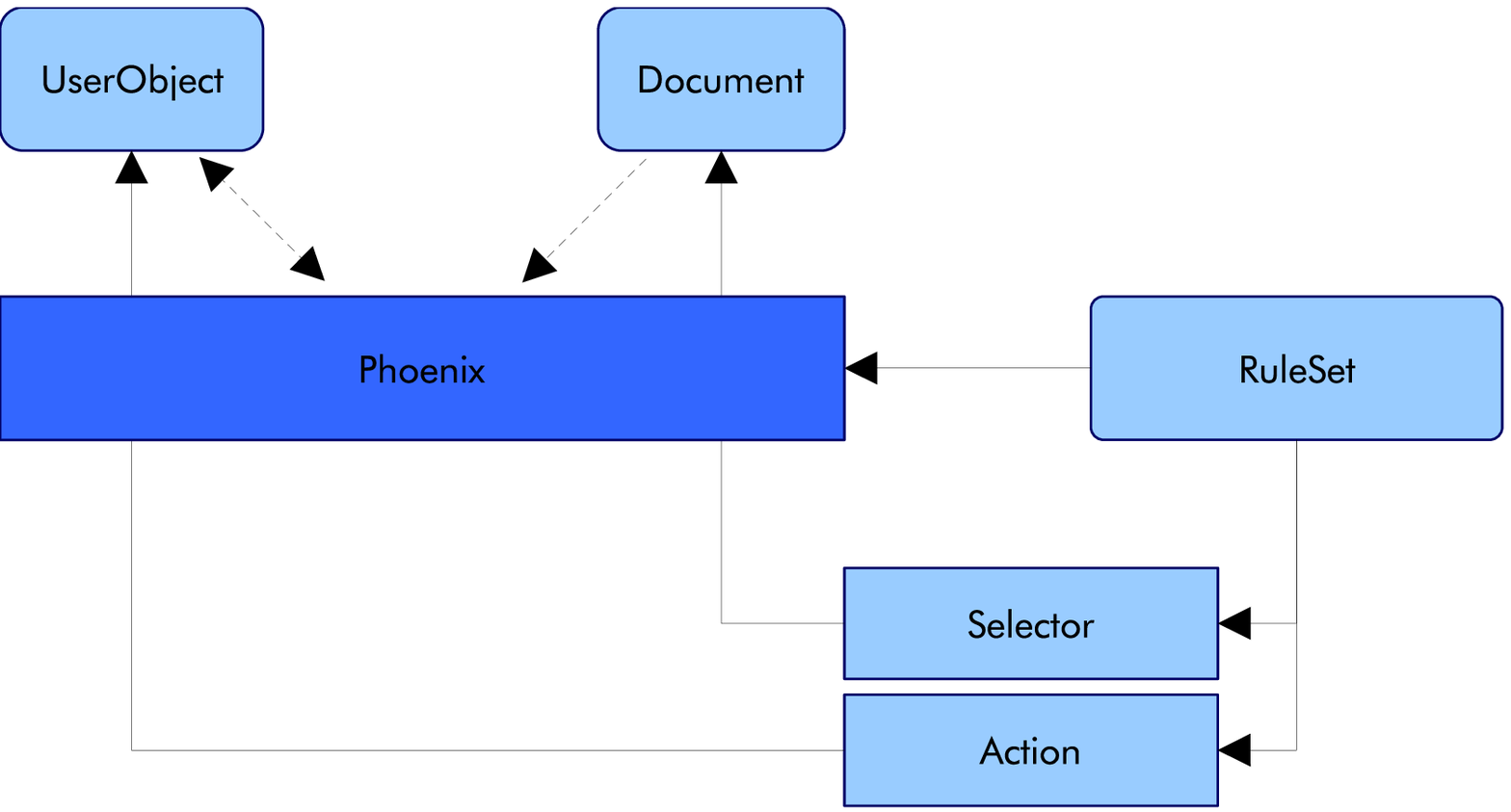} \caption{\label{fig:structure}Phoenix architecture.}
\end{figure}

Phoenix processes arbitrary XML documents as input data. However,
Phoenix' main purpose is to process documents in Star Office/Open
Office or MS Word format, e.g. discharge letters used for case
authoring. Star Office/Open Office .sxw documents essentially are
zipped XML documents, so Phoenix accesses these very easily. Phoenix
is to natively support the Open Document Standard in future
releases.\footnote{OpenOffice document format is basis for the
\textit{Open Document Format for Office Applications}, an OASIS
standard supported by StarOffice, OpenOffice, KWord and hopefully by
future versions of Microsoft Word.} Microsoft Word documents are
converted to .sxw automatically, using Open Office as a conversion
server. Phoenix is also able to process HTML documents by using
JTidy\footnote{\url{http://jtidy.sourceforge.net}} as XML parser.

Phoenix processes XML input documents as DOM\footnote{DOM: Document
Object Model, \url{http://www.w3.org/DOM/}} trees rather than as
character input stream. However, processing does not work on single
DOM tree nodes, but on \textit{blocks} -- node collections specified
by the rule set. A block thus is a document fragment with all
children matching certain criteria. Each rule set defines one or
more \textit{block types}, each specified by an XPath expression, a
starting condition (based upon selectors) and a grouping expression.
Each block type defines a set of rules applied to all blocks of this
type identified in the document. These rules fire on a block, if the
rule's condition (again based on selectors) meets the block's
content. Upon firing, the rules activate an action or recursively
start another rule set on that block.

We decided against XML documents as premium output. Instead, actions
may alter the user object dependent on the block's content. First,
this API-level access to arbitrary user objects enable the use of
pre-existing libraries for knowledge representation, providing
consistency checks, capsulation, and individual persistence. Second,
the more general concept of actions provides a more powerful
processing: Direct generation of XML makes it hard to re-structure
information once written based upon information parsed later on.
However, XML output is supported as a feature (see below) to provide
an easy to use transformation output format.

After Phoenix finished parsing a document, the user object is set up
with all the information from the document. One can now use this
object, e.g. by storing the data to a persistent representation.

\section{Algorithm}

Phoenix starts processing based upon an \code{org.w3c.dom.Node}
representing the input document and a \code{java.lang.Object} as
user object. Utility methods provide transparent access to .sxw,
.doc, and .html documents.

Basis for the parsing process is a \textit{rule set}, providing a
set of \textit{block type} definitions. Each block type is specified
by an \textit{XPath expression}, a \textit{starting condition} and a
\textit{grouping expression}. First, based upon these criteria,
minimal \textit{blocks} are generated by Phoenix: For each block
type, document nodes matching the XPath expression are evaluated
against the corresponding starting condition. If this starting
condition is met, a minimal block is created.

A \textit{condition} either is a terminal condition (one of
\code{Exists}, \code{IntEquals}, \code{TextEquals},
\code{TextContains}, \code{TextStartsWith}, \code{TextEndsWith},
\code{TextMatches}, or \code{Para\-graph\-Start}\footnote{The list
of terminal conditions matches the current needs. It can be extended
easily to reflect future requirements.}) or a non-terminal condition
(one of \textit{and}, \textit{or}, \textit{not}, or
\textit{min-max}). Each terminal condition is configured with a
selector (see \ref{sec:selectors}); some require comparison values,
e.g. \code{TextEquals}. A condition is checked against a block and
returns a Boolean value. This return value is true if and only if
the selector matches the content returned by the selector.
Conditions are not only used for starting conditions, but also for
grouping or end conditions, and action conditions (see below).

In a second phase, the minimal blocks are expanded according to the
corresponding block type's grouping expression. This may be one of
\code{NONE}, \code{GROUPING\_EXPRESSION}, \code{END\_\-EXPRESSION},
and \code{NEXT\_BLOCK}. Both \code{GROUPING\_EXPRESSION} and
\code{END\_EXPRESSION} require an XPath expression.

A \code{NONE}-grouping does not expand the minimal block. These
blocks thus contain a single DOM node. A \code{GROUPING\_EXPRESSION}
adds the sequencing siblings of the block's starting node, if it
matches the XPath specified. Vice versa, \code{END\_EXPRESSION} adds
all siblings as long as they do not match the XPath provided.
\code{NEXT\_BLOCK} as grouping type expands the block up to the
beginning of the next block -- this grouping type is the reason for
the two-phase block construction process.

For each block Phoenix checks all rules defined by this block's
type. \textit{Rules} are condition, action, rule set triples, where
either action or rule set or both may be set. If the condition is
met by the block, the action (see \ref{sec:actions}) fires and the
rule set is applied to this block's content.

These inner rule sets optionally specify pre- and post actions to
switch the user object for the scope of this inner rule set.
Therefore, the pre-action creates and returns a new user object. If
no pre-action is given, the rule set inherits the knowledge
container object. After inner rule set processing is finished, post
action post-processed the extracted information and writes back the
local user object to the original user object.

After Phoenix finished traversing the document, the user object
given is filled with information extracted. The user object then can
be manipulated in arbitrary ways, e.g. stored in a database.

\section{Expansion mechanisms}

To allow Phoenix to fit into multiple environments, it provides
flexible mechanisms for input content selection (selectors) and for
writing result representation (actions). Selectors are to be used in
any conditions (rule set starting conditions, grouping expressions,
and rule conditions) or inside of actions. An action is part of a
rule as defined above.

\subsection{Selectors}
\label{sec:selectors}

Generally, selectors are referenced by their class and must be
implementations of the interface
\code{de.knowit.phoenix.ruleEngine.Selector}. This interface
requires a single \code{get}-method. For a given block, a selector's
\code{get}-method returns an \code{org.w3c.dom.Node}. This Node
usually is a single node or a subset (as
\code{org.w3c.dom.DocumentFragment}) from the block's contents. But,
since one is free to implement arbitrary \code{get}-methods, a
selector might also return information associated to the block, e.g.
style information, or generated information, e.g. Date and Time.

To provide a flexible mechanism, selectors are parameterizable if
they implement the Interface
\code{de.knowit.phoenix.ruleEngine.ParameterizedSelector}. This is
especially useful for the built-in selectors provided by Phoenix
(see table \ref{tab:selectors}). While \code{Identity\-Selector} and
\code{PositionSelector} do not require parameters,
\code{XPathSelector} requires an XPath expression (`xpath') and
\code{RegExpSelector} requires a regular expression pattern
(`regexp') as parameter.

\begin{table}[!ht]
\begin{tabularx}{\textwidth}{llX}
\toprule
\textbf{Class}\footnote{In package \code{de.knowit.phoenix.selectors}} & \textbf{Parameters} & Description \\
\midrule
\code{IdentitySelector} &-& Returns the blocks content as \code{DocumentFragment}\\
\code{PositionSelector} &-& Returns the blocks position in the list of all blocks.\\
\code{XPathSelector} & xpath & Returns the first node matching the XPath expression.\\
\code{RegExpSelector}& regexp & Returns the DOM subtree for witch the text matches the given regular expression. Text nodes might be splitted, structure is cloned as necessary to keep text.\\
\bottomrule
\end{tabularx}
\caption{\label{tab:selectors}Phoenix's built-in selectors.}
\end{table}

If the return value of a selector only depends upon the input block,
the first computation of the return value can be cached to improve
performance. Phoenix already supports this, if a Selector extends
\code{de.knowit.phoenix.ruleEngine.CachedSelector}, overwriting the
\code{handleGet}-method instead of the \code{get}-method. To improve
management of word processing documents, Phoenix provides
convenience methods to read style information -- either directly
applied to the content or via (inherited) masters. Thus, Selectors
can return content based upon the style information associated to
the content, e.g. bold text ending with a question mark.

\subsection{Actions}
\label{sec:actions}

Like selectors, Actions are referenced by their class
(implementations of \code{de.knowit.phoe\-nix.ruleEngine.Action}),
configured by parameters (if the action class implements
\code{de.knowit.phoenix.ruleEngine.ParameterizedAction}), and a
selector (if \code{de. knowit.phoenix.ruleEngine.ActionWithSelector}
is implemented) as a special kind of parameter.

If a rule is activated, the \code{perform}-method of its action is
called, receiving the actual block and the user object as
parameters. In addition, a logger is given, so that all activities
can be logged using the java logging API.

Besides the \code{Trace}-action
(\code{de.knowit.phoenix.actions.Trace}), which writes information
to standard output and logs it, there is a predefined action for
writing extracted information to a DOM tree. We will focus on this
in the next section.

\section{Store extracted information to XML}

As mentioned above, Phoenix provides the possibility to write
extracted data back to XML again. By that way, one can transform
semi-unstructured office documents to structured data represented in
XML without the need to implement extensions.

For XML data storage, an \code{org.w3c.dom.Node} must be used as
user object, usually an \code{org.w3c.dom.Document}. A rule set pre
action
(\code{de.knowit.phoenix.xmlUser\-Object.DescendNodePreAction}) to
locate a node, and an action
(\code{de.knowit.phoe\-nix.xmlUserObject.SetNodeAction}) to set a
node value are provided by Phoenix.  Both actions require two
parameters: `path' and `overwrite'.

Path is in XPath\textsuperscript{--}, a subclass of XPath: It
denotes a sequence of Nodes, separated by a `/'-character. An
attribute node is characterized by an `@' sign and may only be the
last node in a path, e.g. /organization/person/@id.

\code{DescendNodePreAction} selects the node specified by
\code{path} as new user object. Therefore, it creates new Nodes if
\code{overwrite} is true. Otherwise, existing nodes matching the
XPath\textsuperscript{--} expression are re-used, new nodes are
created as necessary. For \code{DescendNode\-PreAction}, \code{path}
must end in an element reference.

\code{SetNodeAction} also takes a selector -- if no selector is
given, \code{SetNodeAction} acts like an \code{IdentitySelector} was
given. Dependent on the node type of the last path element,
\code{SetNodeAction} performs: If the last path element is an
attribute, this attribute's value is set to the text of the node
returned by the given selector. If the last path element is an
element node, the node returned by the selector is added as child to
that node.

\section{Rule Set Definition}

Phoenix usually is running against many documents sharing a static
common structure. Therefore, each application's grammar changes only
little over time. This allows for manual grammar modeling, providing
maximum flexibility.

Phoenix defines rule sets by XML documents according to the
Phoenix-XMLSchema\footnote{http://ki.informatik.uni-wuerzburg.de/\~{}betz/phoenix/}.
Each rule set document provides a root node named \code{RuleSet}.
Each \code{RuleSet} has an \code{ID}. Additionally, the RuleSet node
defines necessary namespaces.

{\small \begin{verbatim}
<RuleSet ID="RS:default"
  xmlns:office="http://openoffice.org/2000/office"
  [...]
  xmlns:text="http://openoffice.org/2000/text"
  xmlns:xsi="http://www.w3.org/2001/XMLSchema-instance"
  xsi:noNamespaceSchemaLocation
    ="http://ki.informatik.uni-wuerzburg.de/
          ~betz/phoenix/phoenix.xsd">
\end{verbatim}}

Furthermore, an inner RuleSet-Node may contain \code{pre}- and
\code{post}-attributes, specifying the pre and post rule set actions
by class name.

{\small \begin{verbatim}
<RuleSet ID="BlockSequence"
  pre="de.d3web.caseParser.actions.examinations.StartCaseParagraph"
  post="de.d3web.caseParser.actions.examinations.EndCaseParagraph">
\end{verbatim}}

Each RuleSet-Node comprises any number of block type definitions,
where each Block has an ID-attribute, a Definition and a list of
Rules.

{\small \begin{verbatim}
<Block ID="body">
  <Definition>
    <Start matches="/text:p | /text:h"/>
    <Condition type="and">
      <Condition type="paragraphStart"/>
      <Condition type="exists"
          selector="de.knowit.phoenix.selectors.RegexpSelector"
          selectorParameters="regexp=\s*(.*)\s*:"/>
    <Grouping type="END_EXPRESSION">
      <GroupingExpression
          matches="descendant-or-self::*[contains(text(),'Ende')]"/>
    </Grouping>
  </Definition>
  <Rules>
    <Rule>
      ...
    </Rule>
  </Rules>
</Block>
\end{verbatim}}

The \code{matches} attribute to start specifies the XPath expression
for the block starting node. The \code{condition} is given by
\code{type}, \code{selector} (optional) and
\code{selectorParameter}s (optional). Conditions may be nested using
the aggregation conditions (\code{and}, \code{or}, \code{not},
\code{min\-max}).

Grouping may be one of the types specified above, where \code{NONE}
is expressed by omitting the \code{Grouping} tag. Thus, the
following grouping tags are valid:

{\small \begin{verbatim}
<Grouping type="GROUPING_EXPRESSION">
  <GroupingExpression
     matches="descendant-or-self::*[contains(text(),'@')]"/>
</Grouping>
\end{verbatim}}

{\small \begin{verbatim}
<Grouping type="END_EXPRESSION">
  <GroupingExpression
     matches="descendant-or-self::*[contains(text(),'Ende')]"/>
</Grouping>
\end{verbatim}}

{\small \begin{verbatim}
<Grouping type="NEXT_BLOCK"/>
\end{verbatim}}

Like the items above, rules are attributed with an \code{ID}. They
share the syntax of \code{Condition} given above. Each rule
possesses at most one \code{Action} tag, specifying the action
\code{class}, and one \code{RuleSet}.

{\small \begin{verbatim}
<Rules>
  <Rule ID="R1">
    <Condition type="contains"
        selector="de.d3web.caseParser.selectors.TitleSelector"
        value="Definition"/>
    <Action class="de.knowit.phoenix.xmlUserObject.SetNodeAction"
        parameters="path=@title;overwrite=false">
†     <Source selector="example.selectors.StartingNodeSelector" /> †
</Action>
    <RuleSet ID="RS:R1">
    ...
    </RuleSet>
  </Rule>
</Rules>
\end{verbatim}}

\section{d3web.CaseImporter}

Our main target was to build an application
(``d3web.CaseImporter\footnote{\url{http://www.d3webtrain.de/author/}}'')
for extracting medical training cases from dismissal records (see
\cite{Betz2004}). With this application, we proof the concept of
evolutionary modeling: Authors of medical training cases re-use
existing documents, altering and extending content as needed.

With only little changes, a dismissal record can be transformed into
a training case: An author needs to make sure that the document's
layout match the requirements given by CaseImporter. He usually
strips unwanted formatting like headers and footers. Also, he
ensures headings to be in the correct format: starting a new
paragraph, boldfaced and ended by a colon. We chose the format used
in most dismissal records, so the need for changes in the document
is minimal. The heading for the list of diagnoses must be
`Diagnosen'. The most important step in this first pass is
anonymization: The author must remove any private data, including
dates and locations.

After the author performed these steps, he can upload his document
to d3web.CaseImporter using a web browser. For each case,
CaseImporter provides him with a log of parsing events (indicating
possible problems using `traffic lights') and a dump of case
contents. Also, the author can directly start his case in
d3web.Train.

As students' pre-knowledge and learning goals require, author
extends his case. He adds texts and images for introduction or
conclusion and multiple choice questions. He improves presentation
by adding images (like x-rays, smears or screenshots of lab data
forms) and he subjoins image interpretation tasks. For relating
observations to diagnoses, author labels both with the same
background color.

From these documents, CaseImporter generates a structured
representation based upon d3web's knowledge model: three
terminologies (examinations, diagnoses, and therapies) are
populated. Content and tasks related to these terminologies. For
example, a diagnose selection task requires the learner to select
diagnoses appropriate to a given situation from the terminology.
Feedback then compares this selection to the list of diagnoses from
the terminology given by the author, respecting even hierarchical
relations.

To implement CaseImporter, we developed appropriate selectors,
actions, and a rule-set. We used selectors basically as shortcut to
simplify rule-set definition and to implement the caching mechanism
outlined above: e.g. \code{TitleSelector} and \code{ContentSelector}
seperate title and content of a paragraph. Actions write to d3web's
\code{CaseObject} and sub-parts, as inner rule sets create and
select appropriate user objects (like \code{CaseParagraph}s). Only
\code{ImageExtraction} action extracts an image included in the
document to the file system, clipping the picture as necessary.

\section{Conclusion}
Since Phoenix as a general purpose tool, it is already in use in
several projects: As a spin-off project from the case extraction
engine, Phoenix parser was integrated into the knowledge modeling
environment \textit{KnowME} to import terminology from text or
document files. The knowledge bases created with KnowME are used
either in d3web\footnote{\url{http://www.d3web.de}} applications or
in the consultation system \textit{Assist}\footnote{Assist by
knowIT-Software GmbH, \url{http://www.knowit-software.de}}.

Completely separated from our main project, Phoenix is also used to
populate a juridical eLearning environment from Word documents.

Future releases of Phoenix will include actions for building
Semantic Web ontologies, building on the Jena
\footnote{\url{http://jena.sourceforge.net/}} framework. By this, we
will carrying on the evolutionary approach to arbitrary Semantic Web
applications, widening the modeling bottleneck.

Experiences show that d3web.CaseImporter matches that goal for
medical case based training systems: It was possible to reduce the
time for settling-in from months down to an hour. Also, time for
developing a single case was reduced, especially when compared to
previous approaches to first build a complete diagnostic knowledge
base for the domain or to reuse an existing one
\cite{Reinhardt1999Diss}.

This speed-up led to an increasing acceptance of case-based training
systems by authors. Now, even inexperienced authors are able to
develop high-quality cases in a reasonable amount of time, e.g. when
preparing a lecture. Training systems built using CaseImporter are
well accepted by students \cite{Kraemer2005}.

As Kraemer, co-author of an onkological system, puts it: ``The
d3web.Train system offers a new and great tool for creating a
training program in a reasonable amount of time''
\cite{Kraemer2005}.


{\ifthenelse{\boolean{publ}}{\footnotesize}{\small}
 \bibliographystyle{bmc_article}  
  \bibliography{cbetz_phoenix} }     


\ifthenelse{\boolean{publ}}{\end{multicols}}{}


\end{bmcformat}
\end{document}